# The Open Language Archives Community:
# An Infrastructure for Distributed Archiving of Language Resources

**Gary Simons**
SIL International, USA

**Steven Bird**
University of Pennsylvania, USA;
University of Melbourne, Australia

## Abstract

New ways of documenting and describing language via electronic media coupled with new ways of distributing the results via the World-Wide Web offer a degree of access to language resources that is unparalleled in history. At the same time, the proliferation of approaches to using these new technologies is causing serious problems relating to resource discovery and resource creation. This article describes the infrastructure that the Open Language Archives Community (OLAC) has built in order to address these problems. Its technical and usage infrastructures address problems of resource discovery by constructing a single virtual library of distributed resources. Its governance infrastructure addresses problems of resource creation by providing a mechanism through which the language-resource community can express its consensus on recommended best practices.

## 1. Introduction

A current trend in literary and linguistic computing is the explosion of digital resources. These resources include not only data (i.e. electronic editions of primary sources and secondary analyses or descriptions), but also the software tools used to create and view electronic data and the how-to documents that give advice about making best use of the data and tools. It is clear that these new electronic media in conjunction with distribution via the World-Wide Web offer a degree of access to resources that is unparalleled in history.

But there is a gap between what users need and what they can achieve today. For instance, even though the electronic resource that a potential user needs may exist, it may not be indexed by search engines. Even if a needed resource has been indexed by a search engine, it will remain inaccessible if the user's search terms do not match the terms by which it was indexed. And even when the user accesses a data resource of relevance, it may remain unusable for want of the tools needed to view and manipulate it or appropriate advice on how to use them. These are all problems of *resource discovery.*

Another major problem area is that of *resource creation.* The proliferation of formats and approaches that have been used by different data providers confounds the average researcher who would like to prepare materials for publication on the web. With the proliferation of approaches, the development of tools for resource creation has been diffused in many directions, with the result that no single approach offers a complete easy-to-use tool set. Most disturbing of all is the fact that the ephemeral nature of the competing approaches (due to the short lifespan of the hardware and software they are based on) is putting these new resources at risk. Unless steps are taken to ensure the longevity of electronic information resources, many of today's resources will be virtually unusable within ten years (Bird and Simons, 2002). The promise of unparalleled access could instead become a reality of unparalleled confusion.

A new direction for humanities computing would be for the community to organize its efforts so as to bridge this gap between present reality and the needs of the community. This paper describes what one subcommunity, namely, those working with language-related resources, is doing in pursuit of this goal. The Open Language Archives Community (OLAC) was founded in December 2000 when a group of nearly 100 linguists, archivists, and software developers gathered in a workshop on web-based language documentation and description. After reaching consensus on the requirements for language-resource archiving (Simons and Bird, 2000a) and on a vision for how acting in community could serve to bridge the gap between the present reality and the envisioned future (Simons and Bird, 2000b), OLAC was launched with the following purpose statement:

> OLAC, the Open Language Archives Community, is an international partnership of institutions and individuals who are creating a worldwide virtual library of language resources by: (1) developing consensus on best current practice for the digital archiving of language resources, and (2) developing a network of interoperating repositories and services for housing and accessing such resources.

This community involves both people and machines in cooperation. This paper describes the infrastructure that has been developed in order to support the needed cooperation. Section 2 describes the technical infrastructure that defines how participating machines interact with other participating machines. Section 3 describes the usage infrastructure that defines how participating people interact with participating machines. Finally, Section 4 describes the governance infrastructure that defines how participating people interact with each other.

## 2. Technical infrastructure

One of the basic requirements of the language resources community is that each sponsoring institution must be able to host its materials on its own web site. The technical infrastructure for OLAC is thus aimed at the problem of resource discovery within a distributed system; that is, "How can a user find relevant resources when those resources are hosted on a variety of web sites?" This is one of the fundamental problems being addressed by the digital libraries community. OLAC is built on a successful resource-discovery infrastructure that was developed within the Digital Library Federation by the Open Archives Initiative (or OAI, see: http://www.openarchives.org).

In a traditional library, the card catalog is the primary tool for resource discovery. In a card catalog, the resources held by a library are described on index cards with enough information to allow users to judge whether a resource seems relevant. Furthermore, each card gives a unique identifier for the resource (typically, a call number) that allows the user to find the resource itself within the library's physical collection.

In the digital analog to the card catalog, the catalog becomes a database. Each record in the database holds the description of a resource. The catalog description of a resource is called *metadata* (i.e. 'data about data'). The particular problem of the technical infrastructure for machine-to-machine interaction in a distributed digital library is to devise: (1) a standard for the representation of metadata (so that all participating institutions will have compatible metadata), and (2) a method of sharing metadata between systems (so that the metadata from all participating institutions may be pooled to form a union catalog for a single virtual library).

In the OAI infrastructure, the metadata standard is an XML representation of the Dublin Core metadata set (DCMI, 1999). The Dublin Core metadata set defines fifteen elements for describing a resource that are both optional and repeatable in any one resource description: Contributor, Coverage, Creator, Date, Description, Format, Identifier, Language, Publisher, Relation, Rights, Source, Subject, Title, and Type.

The OLAC metadata set (Simons and Bird, 2001a) takes these fifteen elements as a base and adds a few more elements to meet the specific needs of the language-resources community. Two elements are added to aid in the discovery of language-related data resources. Whereas the Language element identifies the language in which the resource is written or spoken, a Subject.language element is added to identify the language which a resource is about. Whereas the Type element identifies what type of thing the resource is generically (such as sound or image or readable text), a Type.linguistic element is added to identify what type of thing the resource is from a linguistic point of view (such as a text or a lexicon or a grammatical description). For instance, the following is a metadata description in OLAC format of an electronic dictionary:

```
<olac xmlns="http://www.language-archives.org/OLAC/0.4/">
   <title>Limbu-English Dictionary</title>
   <creator>Michailovsky, Boyd</creator>
   <date code="2002-05-22" refine="modified"/>
   <description>The XML source for a dictionary of the Limbu
```

```
        language of Nepal, consisting of approximately 2,000
        entries. (Size: 1.2M)</description>
    <format code="text/xml"/>
    <publisher>LACITO Project, Centre National de la Recherche
        Scientifique (CNRS)</publisher>
    <language code="en"/>
    <subject.language code="x-sil-LIF"/>
    <type code="Text"/>
    <type.linguistic code="lexicon/dictionary"/>
    <identifier>http://lacito.archivage.vjf.cnrs.fr/archives/
        Nepal/Limbu/dicoLimbu.xml</identifier>
</olac>
```

In order to improve recall and precision when searching for resources, the OLAC metadata standard defines a number of controlled vocabularies for descriptor terms. These appear in the example above as the values of the *code* and *refine* attributes. The most important of these for the language resources community is a standard for identifying languages (Simons, 2000). Language names are not a reliable descriptor for resource discovery since most languages are known by a number of alternate names. Thus OLAC has standardized on using the three-letter codes that SIL International has developed for uniquely identifying the 7,148 living and recently extinct languages listed in the *Ethnologue* (SIL, 2002b). These are augmented with around 300 codes for ancient and constructed languages that are maintained by Linguist List (Aristar, 2002), and 140 two-letter codes from ISO 639-1 that are not ambiguous (SIL, 2002a). For a discussion of the other OLAC controlled vocabularies see Bird and Simons (2001).

The last element in the metadata record above, Identifier, is used to give the URL for the resource when it is available online. This means that the potential users are just one click away from the resource itself when they read its metadata in a catalog search result. However, it is not a requirement that the described resource be available online; the Identifier element, like all elements of the metadata set, is optional in any given metadata record. Thus the OLAC metadata set may also be used to describe the nondigital holdings of a conventional archive. By participating in OLAC and indexing their holdings against specific language identification codes, such archives have the opportunity of alerting people interested in a specific language of their relevant holdings.

The second major problem solved by the technical infrastructure is that of sharing metadata so that a single, pooled catalog of the distributed resources can be constructed. The OAI infrastructure uses a "pull" strategy in which a *service provider* (a site that wants to provide a service based on a pooled collection of metadata records) harvests the metadata from each *data provider* (a participating institution that publishes a repository of metadata records describing the holdings of its archive). In the OAI protocol for metadata harvesting (Van de Sompel and Lagoze, 2001), service providers implement a metadata harvester which requests information from data providers by means of HTTP requests. The protocol defines six possible requests: GetRecord, Identify, ListIdentifiers,

ListMetadataFormats, ListRecords, and ListSets. Data providers implement an interface that decodes the request, queries a local database to retrieve the requested information, and then answers the request by returning an XML document that conforms to a schema specified by the protocol. The service provider in turn parses the returned XML document to extract the desired information and insert it into a database local to its machine. Figure 1 illustrates the basic OAI metadata harvesting model. The circles represent processes and the cylinders represent databases.

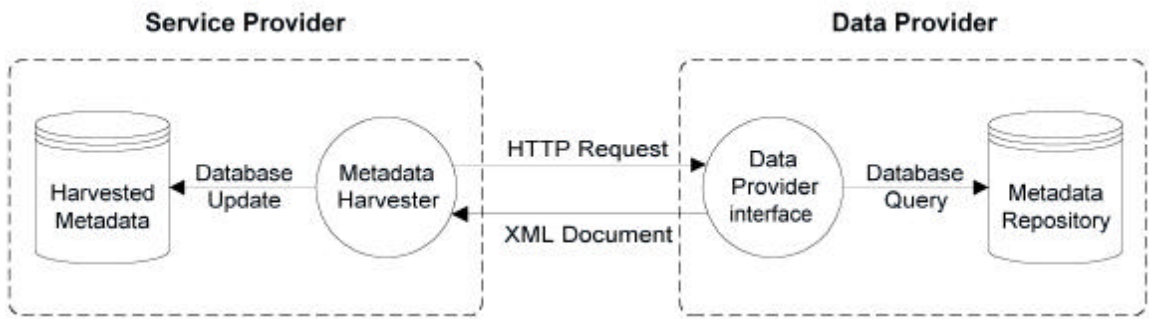

**Fig. 1** The OAI model for metadata harvesting

The OLAC protocol for metadata harvesting (Simons and Bird, 2001c) builds on the OAI protocol. It simply adds a few requirements to the OAI protocol. The chief among these are that the data provider must return metadata records that conform to the OLAC metadata format and that the answer to the Identify request must contain an archive description that conforms to a schema defined by OLAC.

The protocol uses the standard CGI syntax for implementing dynamic web pages. To implement an OAI data provider, the participating archive implements a CGI interface to the local database that contains the archive catalog (using their preferred approach for implementing dynamic web pages, e.g. Perl, JSP, PHP, ASP). The interface program is posted on the institution's web site at a publicly accessible URL. The protocol requests are specified by means of the *verb* parameter appended to the URL. For instance, the following URL issues the Identify request on the data provider for SIL International's *Ethnologue:*

```
http://www.ethnologue.com/oai_server.asp?verb=Identify
```

The interested reader may type this URL into a web browser to see the XML document that is returned. Many of the protocol requests involve additional parameters. For instance, the following parameter string (appended to the same base URL) is used to request the metadata record in OLAC format for the archived item that has the unique identifier of *oai:ethnologue:AAA:*

```
?verb=GetRecord&metadataPrefix=olac&identifier=oai:ethnologue:AAA
```

The ListIdentifiers and ListRecords requests have optional parameters *since* and *from* which specify a date and permit service providers to do incremental harvesting of data providers.

The right half of Fig. 2 gives an overview of the OLAC technical infrastructure that defines the machine-to-machine communication within the community. The dashed line that divides the technical infrastructure into two halves represents the division between centralized services on the left and the components distributed at participating archives on the right. In the diagram the arrows represent the flow of information requests; the receiving component typically returns a result in the reverse direction of the arrow. Note that the upper right portion of the diagram incorporates the basic OAI model given above in Fig. 1.

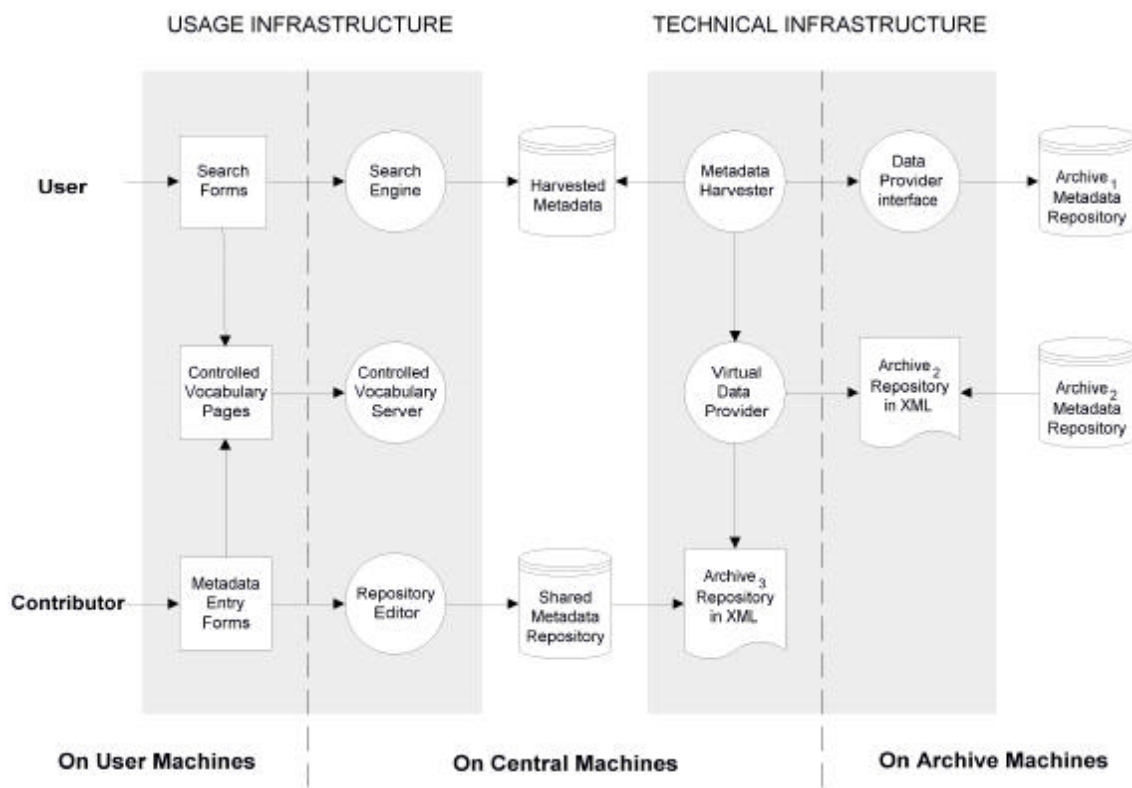

**Fig. 2** Overview of the OLAC usage infrastructure and technical infrastructure

Another bit of technical infrastructure that is needed to allow any site to become a service provider that harvests OLAC metadata is a machine-readable list of every participating archive and the base URL for its data provider. This list is a dynamically generated XML document that a service provider may access at the following URL:

        http://www.language-archives.org/register/archive_list.php4

The means for adding an archive to this list is a matter of usage infrastructure and is covered in the next section.

Implementing a data provider interface is not difficult for a programmer who understands CGI interfaces and dynamic database connections, but many institutions that could be providers of language resources do not have this kind of capability. In order to make it easier for such institutions to participate, OLAC has developed two more pieces of technical infrastructure: Vida (for Virtual Data Provider) and ORyX (for OLAC Repository in XML). ORyX is an XML schema that permits all of the information in a metadata repository, including both identification of the archive and all the metadata records, to be represented in a single XML document. Vida is a process that implements the OAI data provider interface for an ORyX document. For instance, the following is the URL for the ORyX that the Linguistic Data Consortium has posted describing its collection:

```
http://www.ldc.upenn.edu/Projects/OLAC/ldc.xml
```

The interested reader may browse that URL in order to see the format of an ORyX document. Vida is a PHP4 script posted on the OLAC web site at the following URL:

```
http://www.language-archives.org/tools/vida.php4
```

The web address of the ORyX is appended to the Vida URL to form the base URL that a service provider uses to harvest metadata from the ORyX. Enter the Vida URL above in a web browser to see a page that describes how this works and gives an example.

This alternative approach to becoming an OLAC data provider is illustrated in the middle row of the technical infrastructure in Fig. 2. In this case the participating institution (Archive$_2$) chooses to generate a static XML document to represent the information in its catalog database. That XML document is published on a publicly accessible web site, and its URL is appended to the Vida URL when registering the base URL of the archive. When a service provider runs a metadata harvester and encounters such a URL, the Vida process on the central machine receives the request and goes to the ORyX on the archive's machine to extract the information that goes in the dynamically generated XML response to the protocol request.

## 3. Usage infrastructure

The left half of Fig. 2 diagrams the usage infrastructure that has been built to allow the community of people interested in language resources to interact with the machines that provide services for this community.

The most important piece of infrastructure for the user community at large is the search engine that allows any user to search for resources in the combined catalog of metadata harvested from all participating archives. The top left corner of Fig. 2 shows this search engine as a process running on a central machine; it accesses the database of harvested metadata in order to generate results from user requests. Users make search requests through interactive forms that run in a web browser on their own machines. Linguist List, which with over 15,000 subscribers has become the virtual hub of the linguistics community on the web, is hosting the central union catalog service for OLAC. The URL for the catalog search form is:

```
http://www.linguistlist.org/olac
```

Since the data providers and the harvesting protocol are open, any institution is free to implement a metadata harvester and offer a search interface to OLAC resources. In particular, a site with a focus on a specific part of the world or on a specific language family or on a specific kind of data resource could selectively harvest resources that match its area of specialization and offer a service that presents OLAC resources related to its focus.

The purpose of the controlled vocabularies discussed as part of the technical infrastructure is to improve recall and precision in searching. To receive this benefit, users must be able to find the right descriptor in a controlled vocabulary as they are formulating a query. Thus another aspect of the usage infrastructure (shown in the middle left of Fig. 2) is controlled vocabulary servers. These are centralized services that serve web pages documenting the descriptors of the controlled vocabularies. The largest of these is the language identification server hosted by SIL International at `www.ethnologue.com`. Not shown in the diagram is the fact that the catalogers for the metadata repositories on archive machines are also able to consult these servers in order to find the right descriptors during the cataloging process.

Many potential contributors of language resources (whether they be institutions or individuals) do not have the capacity to post metadata on their own machine, whether through a data provider interface or a repository in XML. In order to open OLAC participation to even these potential contributors, another component of the usage infrastructure has been implemented. ORE, for OLAC Repository Editor, is a form-based metadata editor that any potential contributor may run from a web browser. This service, too, is hosted by Linguist List at:

```
http://www.linguistlist.org/ore/
```

This URL invokes a login page; it links to a simple registration form which anyone may use to define a login password. Once logged into ORE, users may enter identification information for their archive and create metadata descriptions of archive holdings. The results for all archives are stored in a shared database on a central computer. When the contributor instructs ORE that the archive's metadata repository is ready to be published, the information in the database is written as a repository in XML. The ORyX is automatically posted at a publicly accessible URL on a central computer and registered as a data provider serviced through Vida. Archive$_3$ in Fig. 2 illustrates this approach to becoming an OLAC data provider.

A final component of the usage infrastructure (not shown in Fig. 2) is the OLAC web site at `www.language-archives.org`. This is where users come to conduct business with the community. It provides current news, the latest versions of all OLAC standards and other community documents, a directory of participants, links to service providers, and resources (including sample code) for implementers of data providers. It also includes forms for subscribing to an OLAC mailing list and for registering a new data provider so that it will be added to the list of harvested archives.

## 4. Governance infrastructure

The governance infrastructure is the aspect of OLAC that supports the interaction among the human participants of the community. It is defined in the OLAC process document (Simons and Bird, 2001b). After summarizing the purpose, vision, and core values of OLAC, the process document does two things: it defines how OLAC is organized and how it operates.

The organization of OLAC is defined in terms of the groups of participants that play key roles. There are coordinators who oversee the operation of the OLAC process and an advisory board of recognized leaders in the language resources arena who advise the coordinators on the concerns of their particular subcommunities and promote OLAC within those subcommunities. The remaining categories of participation—participating archives and services, prospective participants, working groups, and participating individuals—are open to any who want to participate in these ways. For instance, the OLAC home page has a link on "How to become a data provider" (i.e. participating archive) and has a fill-in form for subscribing to the OLAC-General mailing list (which is how to become a participating individual).

It is through documents that OLAC defines itself and the practices that it promotes. Thus the operation of OLAC is defined in terms of a process for generating and ultimately adopting documents. There are two key kinds of documents: *standards* which define the technical and governance infrastructures, and *best practice recommendations* which express the consensus of the community on best current practice for the digital archiving of language resources.

The opening section of this article introduced two major problem areas confronting the language resources community: resource discovery and resource creation. The standards that govern the technical infrastructure lie at the heart of OLAC's contribution toward solving the problems of resource discovery. Best practice recommendations, on the other hand, are the means by which OLAC addresses the problems of resource creation. The central problem for resource creation is the proliferation of approaches and formats. This proliferation leaves resource developers confused as to how to proceed, dilutes the utility of software tools, and puts resources at risk of becoming inaccessible as formats change. The only way to address this problem is for the community to determine which of the available practices seem to best ensure the longevity and portability of language resources (Bird and Simons, 2002) and then to follow these practices.

The OLAC document process offers the language resources community a means by which it can reach consensus on recommended best practice. The process defines how documents are generated and how they progress from one status to the next along the five-phase life cycle of development, proposal, testing, adoption, and retirement. The initial development takes place in the context of a working group. When a working group is formed, a call for participation goes out on the OLAC-General mailing list and any participating individual may join the working group. When the working group has a satisfactory draft of a recommendation, the document achieves proposed status and enters the proposal phase in which all subscribers to OLAC-General are invited to give feedback on the document, including a summary evaluation as to whether it is ready to proceed to the next phase of the process.  When at least 80% of the respondents agree that it is ready, the document achieves candidate status and enters a testing phase during which members of the community attempt to apply the proposed recommendations for a period of months. User feedback leads to final revisions and then a final ballot on the revised best practice statement among the subscribers to OLAC-General. When at least 80% of the respondents agree that the document represents best practice, then the document attains the status of being adopted as recommended best practice.

## 5. Conclusion

During its first year of operation, 2001, the basic infrastructure for OLAC was developed. During the second year, 2002, the focus has been on enlarging the community of participating archives. The standards that define the technical infrastructure have been frozen in candidate status so that member archives need not worry about a moving target as they implement an OLAC data provider. By the time of writing (mid 2002), twenty institutions have published metadata repositories containing a total of around 30,000 records. Based on the experiences of the archives that participate in the first two years, the standards will be refined and formally adopted by the community during the third year, 2003. All individuals and institutions who have language-related resources to share are enthusiastically invited to take part in this new direction for humanities computing that seeks to build consensus on best practices for digital language-resource creation and to build a distributed virtual library to support discovery of resources that document and describe human languages.